%% file: emnlp2020.tex
\documentclass[11pt,a4paper]{article}
\usepackage[hyperref]{eacl2021}
\usepackage[utf8]{inputenc}
\usepackage{times}
\usepackage{latexsym}

\usepackage{pgfplots}
\usepackage{amsmath}
\usepackage{subcaption}
\usepackage[htt]{hyphenat}
\usepackage{booktabs}
\usepackage{multirow,multicol}

\pgfplotsset{
  every axis plot/.append style={line width=0.9pt}
}

\usepackage{microtype}

\aclfinalcopy
\def\aclpaperid{395}

\title{Few-shot learning through contextual data augmentation}

\author{Farid Arthaud~\thanks{~~Work done at University of Edinburgh} \\
  Ecole Normale Supérieure \\
   {\footnotesize \texttt{farid.arthaud@ens.psl.eu}} \\
   \And
  Rachel Bawden~\footnotemark[1] \\
  Inria \\
   {\footnotesize \texttt{rachel.bawden@inria.fr}} \\ \And
  Alexandra Birch \\
  University of Edinburgh \\
   {\footnotesize \texttt{A.Birch@ed.ac.uk}} \\
   }

\begin{document}

\maketitle

\begin{abstract}
Machine translation (MT) models used in industries with constantly changing topics, such as translation or news agencies, need to adapt to new data to maintain their performance over time.
Our aim is to teach a pre-trained MT model to translate previously unseen words accurately, based on very few examples.
We propose (i)~an experimental setup allowing us to simulate novel vocabulary appearing in human-submitted translations, and (ii)~corresponding evaluation metrics to compare our approaches.
We extend a data augmentation approach using a pre-trained language model to create training examples with similar contexts for novel words. 
We compare different fine-tuning and data augmentation approaches and show that adaptation on the scale of one to five examples is possible.
Combining data augmentation with randomly selected training sentences leads to the highest BLEU score and accuracy improvements.
Impressively, with only 1 to 5 examples, our model reports better accuracy scores than a reference system trained with on average 313 parallel examples.
\end{abstract}

\section{Introduction}
News agencies dealing with articles in multiple languages often rely on machine translation (MT) to provide an initial translation, which a journalist reworks into a final article.
This can involve changing the structure or phrasing, but also correcting (or \textit{post-editing}) mistranslated words or expressions, which can frequently occur when new topics emerge, bringing new vocabulary that has been rarely or never seen in the data used for training.
The willingness of a journalist to use MT technology is dependent on the general quality of the models, but also on whether they can learn from the journalist's corrections, to avoid them having to correct the same errors time and time again.

Various strategies have been explored to learn from a journalist's post-edits. One option is to use an automatic post-editing (APE) model trained on the journalist's post-edits.
However, state-of-the-art APE systems \citep{junczys-dowmunt-2017-amu} typically require large numbers of post-edits for training, which are rarely available or hard to generate (particularly for low-resource languages).
An alternative, commonly used strategy is to fine-tune models to in-domain data, but this is prone to overfitting~\cite{miceli-barone-etal-2017-regularization}.
More advanced ways of continually learning through fine-tuning have been explored, selecting similar training instances based on their similarity with test sentences~\citep{li-etal-2018-one,turchi-continuous-2017}.
These methods achieve good results according to automatic MT metrics but can also overfit when training is continued.
So far there has been little focus on the speed of adaptation required
---so as to minimize the number of human interventions required---and the trade-off between successfully adapting to specific new post-edits and maintaining a good global translation quality.

In this article, we aim to explore in more depth this trade-off between overall translation quality and the ability of the MT model to learn specific novel words in the context of life-long learning of MT from journalistic post-edits.
We explore the setup in a simulated few-shot learning scenario, whereby we track the translation performance on specific test words that are removed from the training data and are gradually re-introduced through fine-tuning. To improve the model's capacity to learn these new examples, we explore an approach similar to~\citet{turchi-continuous-2017} and inspired by a contextual data augmentation technique used for rare word translation \citep{fadaee-etal-2017-data} to reduce the number of corrections that need to be seen.

We show in our experiments on the Gujarati-English language pair that it is not only possible to surpass the accuracy of our baseline fine-tuning approach, but also of a reference model which has already seen the new words dozens to hundreds of times during training. 
However we find that in most cases adaptation to new words comes at a variable cost in BLEU \citep{papineni-etal-2002-bleu}, due to overfitting to the new examples.
We show that this cost can be kept at a minimum by padding our data-augmentation approach with randomly selected sentences from our training set. The appropriate choice of hyper-parameters is also important for final performance.

Our code is freely available online.\footnote{\url{https://gitlab.com/farid-fari/fewshot-learning}}

\section{Related work}
The topic of few-shot learning from post-edits is relatively novel, and we were therefore left with few comparison points.
A somewhat similar task that requires quick adaptation is low-resource MT, for which transfer learning~\cite{zoph-etal-2016-transfer} and meta-learning~\cite{gu-etal-2018-meta} approaches exist.
These techniques generally apply for adaptation from hundreds of thousands of sentences, rather than a dozen available in our scenario; this is because we aim to learn individual new words rather than a whole language or domain.

A widespread technique in the MT literature to adapt a model to new data is fine-tuning, often used for domain adaptation.
\citet{turchi-continuous-2017} and \citet{li-etal-2018-one} explore the use of a similarity search in the training corpus in order to fine-tune an MT model before translating a novel sentence or to gradually adapt a model to post-edits.
This approach does not apply to our scenario, where new words appear and therefore a similarity search cannot yield sentences containing these new words.
Our baseline approach (which we call \texttt{finetune}) is present in these works, known as \textit{adaptation a posteriori} in~\citep{turchi-continuous-2017} and also appears as \textit{single-sentence adaptation} in~\citep{kothur-etal-2018-document}.
A challenge these works report with fine-tuning is overfitting, which we encounter systematically when evaluating our various techniques.
Works on fine-tuning also explore several regularization techniques~\cite{simianer-etal-2019-measuring} when adapting to new data, which we choose to leave out of our comparison due to added hyper-parameter choices and complexity of implementation in our experiments -- we do however believe that future work implementing these techniques could potentially outperform ours.

\citet{fadaee-etal-2017-data,kobayashi-2018-contextual,wu2019conditional} and \citet{gao-etal-2019-soft} explore a similar contextual data augmentation technique, albeit in different scenarios and with different goals.
This technique synthesizes new sentences by using sentences from the training set and substituting different words into them.
In \citep{fadaee-etal-2017-data} the goal is to enhance overall translation performance by focusing on words that appear rarely in the training data, but in our case we are training our system to learn new words which were not in the training set at all, based simply on a ground-truth translation of this new word by a human.
Moreover, our technique uses more recent tools and techniques such as the BERT contextual language model~\cite{devlin-etal-2019-bert}.
\citet{kobayashi-2018-contextual} and \citet{wu2019conditional} also use BERT for contextual data augmentation, but with a goal of improving language model tasks such as sentiment analysis.
The constraints for this task are very different; rather than having to produce a translation for augmented data, these approaches have to maintain the sentiment label of the sentences.
\citet{gao-etal-2019-soft} work in a similar context to us, but focus on overall translation performance rather than learning new words, and apply contextual data augmentation during the training step, thus removing the challenge of adapting to new data as it becomes available.

Similar to us, \citet{huck-etal-2019-better} focus on improving the MT of words which are unseen in the training set.
They use bilingual lexicons to hypothesize translations for their unseen terms.
They find these translations in monolingual target side data and backtranslate them inserting the unseen term.
They show that this improves translation performance in the medical domain. However they do not analyze the accuracy of translation of the novel terms, or explore how fast you can learn from very few examples.

\section{Lifelong learning from post-edits}
MT models are inevitably adapted towards the topics and vocabulary from the time period 
associated with their training data.
In the long term, they therefore struggle to correctly translate novel words and expressions associated with new topics, unless they can be adapted to them.
This is particularly a problem in journalism, where current topics and names are constantly changing. A prime example is the recent COVID-19 pandemic: prior to January 2020, newspapers would contain little to no mention of the words \textit{coronavirus}, \textit{respirators}, \textit{PPE masks} and \textit{hydroxychloroquine}.
On top of these topic-specific words, there may also be novel but very frequent expressions that are initially hard to translate, such as \textit{flattening the curve} and \textit{social distancing}, which are likely to be poorly translated and therefore need to be subsequently corrected by the journalist.

We are interested in developing approaches to quickly learn from journalistic post-edits in a way that maintains the general high translation quality of the model.
In order to analyze and objectively evaluate our different approaches, we simulate this scenario in a reproducible way by using a publicly available corpus in which we select the rarest words and separate out all sentences containing them: these words will be our \textit{evaluation words}.
These words must appear at least a few times in the test set as well in order to enable proper evaluation of adaptation using our approaches.
The sentences separated from the training set containing our evaluation words will be used to simulate reference sentences submitted by a journalist for our models to use for adaptation.
Finally, the training set with the evaluation words removed will be our \textit{filtered training set}, used for initial training of our model.

Our test set is also made up of publicly available data sets whose sentences also contain the evaluation words we selected.
This setup prohibits the use of a development set and a test set due to the dependency of the rare words on the test set: they must appear a minimum number of times in the test set to have a way of evaluating our approaches.
In turn, the choice of rare words changes the training set and therefore the models we train.
We are therefore in a transductive learning scenario, where our method is adapted to the task we aim to solve.

We also use as a reference point a model which has been trained on the complete (unfiltered) training set, which is a very strong comparison point since it has seen many more occurrences of the rare words we selected than our few-shot models.

% EXAMPLE FIGURE
\begin{figure*}[ht]
\centering
\includegraphics[width=\linewidth]{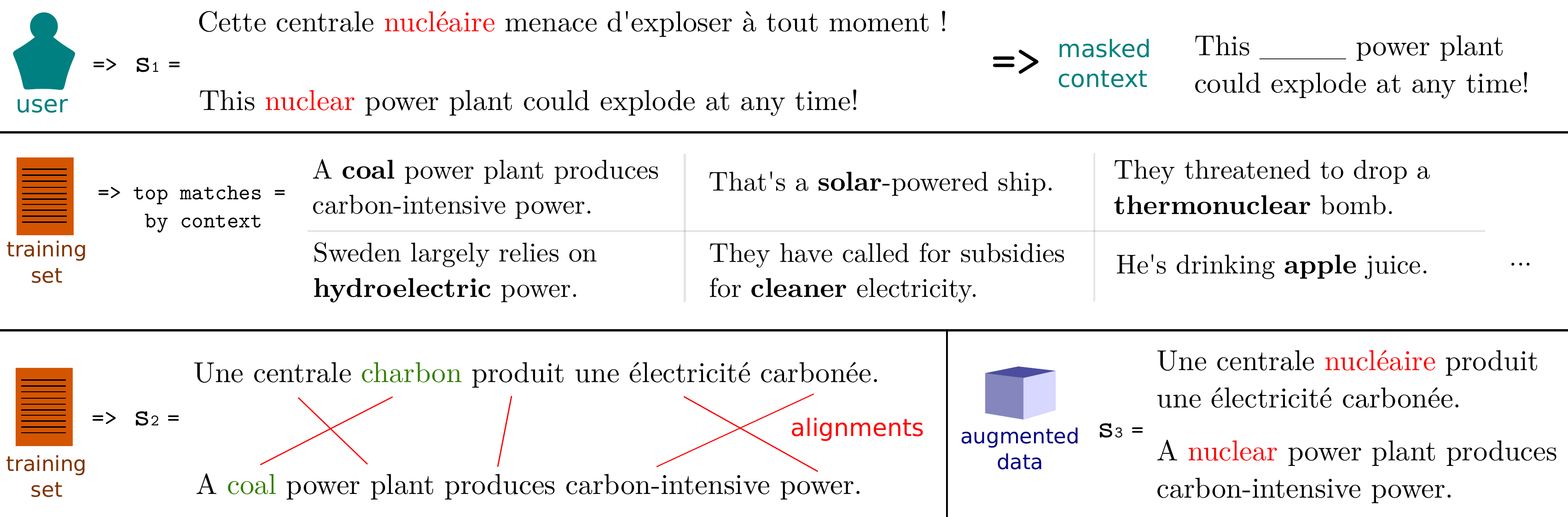}
\caption{Contextual data augmentation applied to an example sentence}
\label{fig:example}
\end{figure*}

\section{Our approaches}
We propose and evaluate four approaches:
\texttt{finetune} (our baseline), \texttt{randompad}, \texttt{augmented} and \texttt{half}.
The first two approaches only involve fine-tuning whereas the latter two include contextual data augmentation.
Each approach works from a given number of reference sentence pairs containing our evaluation words, acting as human-provided references.
In our experimental setup, these sentences are chosen randomly among the held-out sentences from the filtered training set.

\subsection{Baseline: fine-tuning}
Our first approach, which is also our baseline, is naive fine-tuning referred to as \textit{adaptation a posteriori} in~\citep{turchi-continuous-2017}, and which we refer to as \texttt{finetune}.
When presented with a set of reference sentence pairs to adapt to, our model is trained from its previous state for a few epochs using the sentence pairs.
This approach is especially prone to overfitting as it will see very few sentences, and they all contain a potentially repetitive set of words.
To minimize this effect, we choose to fine-tune on a batch of several reference sentence pairs rather than individually for each evaluation word, so as to diversify the data fed to the model.

\subsection{Padded fine-tuning}
The second approach, \texttt{randompad}, attempts to minimize the overfitting from \texttt{finetune} by introducing variety and generality to the fine-tuning set.
This is done by adding randomly chosen sentences from the filtered training set to the reference sentences, which is known as \textit{padding}.
These random sentences do not contain the evaluation words and so encourage generalization during fine-tuning.

\subsection{Contextual data augmentation}
The aim of this approach, similar to the one used by \citet{fadaee-etal-2017-data} and which we refer to as \texttt{augmented}, is to use the reference sentences provided by the journalist to create additional novel sentences containing the evaluation words.
If successful, this approach simulates the presence of more reference sentences containing the new words, which accelerates learning, effectively reducing the number of corrections required from the human translator or journalist.
More formally, given a reference sentence pair \( ( s_{\textit{src}}, s_{\textit{tgt}}) \) containing an unseen word \( w \in s_{\textit{tgt}} \), the goal is to generate new sentence pairs containing \( w \) in the target language, varied enough to be beneficial in fine-tuning.

Our data augmentation process has 2 steps: (i)~for each reference sentence pair provided by the journalist containing a word of interest $w$, find sentences from the training corpus with similar contexts to the one surrounding $w$, and then (ii)~insert $w$ and its translation into the retrieved sentences.

We illustrate this process in Figure~\ref{fig:example}, using the French-English language pair to aid readability: the sentence pair \( s_1 \) containing the new word \textit{nuclear} is provided to our system by the journalist or translator.
We first find sentences in the training set that contain a word (in bold) with a similar context to that of $w$ in  $s_{\textit{tgt}}$.
In each of these sentences, the word in bold could be replaced by \textit{nuclear}, although there are sometimes false positives such as the last sentence: \textit{He's drinking \textbf{nuclear} juice}, would not be a good reference sentence for training.
Next, we use an alignment tool to replace the aligned source word (e.g.~\textit{charbon}) in the selected sentence pair \( s_2 \) with the word \textit{nucléaire}, providing a brand new sentence pair \( s_3 \) for the reference word \textit{nucléaire}-\textit{nuclear}.

\paragraph{Finding similar contexts}
The first step is to find suitable sentences in which we can substitute \( w \) and its translation.
As in \citep{wu2019conditional}, we use the BERT contextual language model \citep{devlin-etal-2019-bert} to provide a contextual representation of $w$ in $s_{\textit{tgt}}$, noted $v$.
This feature vector is taken from the second-to-last layer of BERT (usually used to compute vocabulary probabilities through a softmax layer).
Given the masked and bidirectional nature of BERT, this contextual representation contains information about the context surrounding \( w \) in \( s_\textit{tgt} \), with no prior knowledge of \( w \) required by BERT.
The advantage of this approach is that truly novel words can be data-augmented using this technique with any pre-trained BERT model.

The context search consists in
%\begin{enumerate}
    %\item 
    (i)~extracting the feature vector \( v \) for \( w \) in \( s_\textit{tgt} \),
    %\item 
    (ii)~for each sentence in the filtered training set, randomly selecting a word $u$ in it and computing the corresponding feature vector\footnote{Comparing all positions in all sentences would lead to the best results, but is inefficient. With sufficient training examples (as is the case here), randomly choosing a single position in each sentence avoids having to to do all computations and still provides varied contexts that can be tested.} and
    %\item 
    (iii)~selecting the top \( k \) sentences based on the cosine similarity of these feature vectors with \( v \).
%\end{enumerate}

This is shown in Figure~\ref{fig:example} where 6 sentences have been selected based on their feature vector similarity to \( s_1 \).
These sentences are selected because the context surrounding the random word chosen in each of them is similar to the context surrounding \( w \) in \( s_1 \), which generally means the bold words loosely correspond to adjectives describing power plants.
Using a masked model means that knowledge of the meaning of \( w \) is not required, but only an understanding of the context in \( s_{tgt} \).
An additional real use example is provided in Appendix~\ref{app:examples}.

This process significantly differs from~\citet{fadaee-etal-2017-data}, which iterates over the training set to find sentences for which the language model gives a high probability of the rare words appearing.
Moreover, by using feature vectors rather than raw probability distributions (as in previous works), we capture richer information about the surrounding sentence than only which words are likely to replace \( w \).

\paragraph{Creation of new training examples}
The second step is to substitute the randomly masked word \( u \)---which is \textit{coal} in our example---in the retained sentence pair \( s_2 \) with \( w \) in the target language and its translation in the source language.
In the target language, we simply replace \( u \) by \( w \), but the task is harder in the source language since we have no prior information as to which word translates to \( u \), or which word translates to \( w \) to replace it with.
In our scenario, we assume that the translation in the source language of \( w \), noted \( w' \), is known to us -- for example, a human translator providing \( (s_\textit{src}, s_\textit{tgt}) \) could identify \( w' \) in \( s_\textit{src} \).
Having no human translator for our experiments, we train an alignment tool on the complete unfiltered training set and use it to select the words most often aligned to each \( w \) from our evaluation words.
A different approach could be used in future work, by training the alignment tool on the filtered training set and then re-training it on reference sentences as they are provided, removing the need for human intervention or prior knowledge of \( w \).
Note that all alignment steps rely on automatic tools which can introduce some noise in the process. However, we observed satisfactory results, and this allows for the entire process to be automated.

Once the translation \( w' \) for \( w \) is established, we have to determine which word or words in the source sentence must be replaced by \( w' \).
This is done by using an alignment tool trained only on the filtered training set, to compute the aligned word to \( u \).
The word aligned to \( u \) is replaced by \( w' \); if multiple consecutive words are aligned to \( u \) they are all replaced with \( w' \), and if no words or non-consecutive words are aligned to \( u \) then the sentence is discarded.
This is illustrated in Figure~\ref{fig:example}, where  \textit{coal} is successfully aligned to \textit{charbon}, allowing us to replace it with \textit{nucléaire}.

The final result in the example is \( s_3 \), a sentence pair which makes sense and will be useful for the translation of \textit{nucléaire} to \textit{nuclear}.
The sentences generated by this method will at the very least contain \( w' \) and \( w \) in roughly aligned source and target positions
and the best sentences will have the same quality as human-provided references.

\subsection{Padded data augmentation}
The \texttt{augmented} approach does not benefit from the added generality of the \texttt{randompad} approach; it could suffer from overfitting due to the repetition of our evaluation words (and their semantic fields) in the fine-tuning sentences.
This is addressed by the \texttt{half} approach, designed to overcome potential overfitting in the \texttt{augmented} approach.
This approach uses half of the synthesized sentences from \texttt{augmented} and replaces the other half with random sentences from the filtered training set.
This provides more sentences containing our rare words through augmentation but also more generality from the random sentences, aiming to combine the strengths of both approaches.

\section{Experiments}
\subsection{Experimental setup}
We use the low-resource Gujarati to English language direction.
Our base model is the same as the University of Edinburgh's submission to WMT 2019~\cite{bawden-etal-2019-university}, except that its training set is filtered to remove sentences containing words previously selected for evaluation.
The model is a base transformer model~\cite{vaswani2017attention} with 6 encoder and decoder layers,  feed-forward dimension 512, 8 transformer heads, and dropout of \( 0.1 \). We train using the Marian toolkit \citep{junczys-dowmunt-etal-2018-marian} and the Adam optimizer \citep{kingma2014adam}.\footnote{We experimented with SGD but observed no significant differences with Adam.}
The complete training script with additional parameters can be found on the webpage for the University of Edinburgh's WMT19 submission.\footnote{\url{http://data.statmt.org/wmt19_systems/en-gu/train.sh}}
We use an identical version of this model trained on the full, unfiltered training set as a reference point called \texttt{all-data}.

The main preprocessing, data augmentation and training scripts are freely available online: \url{https://gitlab.com/farid-fari/fewshot-learning}.

We use the pre-trained BERT model from Huggingface~\cite{wolf-etal-2020-transformers}: the \texttt{bert-large-uncased-whole-word-masking} variant, which has the benefit for our use case of masking whole words rather than subwords.

\subsection{Data and preprocessing}
Our training data consists of both genuine Gujarati-English parallel data and backtranslations
from the WMT19 news translation task \citep{barrault-etal-2019-findings}.
The backtranslations are produced as described by \citet{bawden-etal-2019-university} and we follow their method of first training on a mixture of backtranslations and parallel data before fine-tuning on the genuine parallel data only. As mentioned previously, we filter out English words selected for evaluation from both the synthetic and genuine parallel data.
However, only the genuine parallel data is used to select augmentation candidates to ensure their high quality.
For testing, we use the concatenation of \texttt{newstest2019} and \texttt{newsdev2019}.

\paragraph{Preprocessing}
All data augmentation was run after tokenization but before subword segmentation to keep a consistent notion of a `word'.
We first apply tokenization using the Moses tokenizer \citep{koehn-etal-2007-moses} and then apply sub-word segmentation using the BPE strategy~\cite{sennrich-etal-2016-neural} and the \texttt{fastbpe} implementation.\footnote{\url{https://github.com/glample/fastBPE}}$^,$\footnote{We reuse the pre-processing scripts available at \url{http://data.statmt.org/wmt19_systems/scripts/}.}
Word alignment is carried out using
\texttt{fast\_align} \citep{dyer-etal-2013-simple}.\footnote{\url{https://github.com/clab/fast_align}}

\paragraph{Evaluation data}
The rare words to filter out are chosen to be the 100 rarest words in the training set that appear at least 5 times in the test set and 20 times in the training set, and are manually filtered down to 96 words to exclude low quality choices (such as plurals and punctuation).
Their frequency in the unfiltered training set (used by our reference model) ranges from 20 to 775 occurrences, with a mean of 313 and a median of 275 occurrences.
The complete word list can be found in Appendix~\ref{app:filtered}.
These words appear in 701 of the 3,014 test set sentences, meaning that the BLEU score is computed on a majority of sentences not containing our evaluation words, thus giving us a good overview of how general translation performance is affected.
They appear in only 26,910 of the 8.5M training sentences,\footnote{Both genuine and synthetic parallel training sentences.} meaning that the filtered training set is almost the same size as the original full training data.
The genuine parallel data contains 40k sentences which were used for context search in the data augmentation steps.

\paragraph{Evaluation setup}
Our aim is to evaluate how well the model adapts (through fine-tuning of the model) to the gradual addition of reference examples containing the 96 evaluation words, which were absent from the initial training data.
For each evaluation word, we randomly choose 20 reference sentences that contain the word, and for each of the four approaches, we successively make 1, 2, 3, 5, 10, 15 and 20 reference occurrences of each word available to the model to learn from, evaluating at each step.
Everything is run as a batch over all 96 words; for instance in an experiment using 3 occurrences, fine-tuning is conducted over 288 reference sentences, with each reference word occurring 3 times.\footnote{The number of sentences is very slightly lower due to some sentences containing 2 evaluation words, but this is rare enough that it does not meaningfully impact our results.}

Each approach that uses padding and/or augmentation requires choosing a ratio \( r \) between the total number of fine-tuning sentences and the original number of reference sentences provided by the journalist. For a word \( w \), as the number of occurrences offered to the model grows from 1 to 20, the number of fine-tuning sentences grows from \( r \) to \( 20r \).
This ratio $r$ is calculated as follows:
\vspace{-0.15cm}
\begin{align}
n_{total} &=  n_{ref} + n_{synth} + n_{rand} \\
r &= \dfrac{n_{total}}{n_{ref}},
\end{align}
\vspace{-0.35cm}

where $n_{ref}$, $n_{synth}$ and $n_{rand}$ refer respectively to the number of reference, synthetic and random sentences. The chosen ratio is written in brackets following the name of each approach.

For the \texttt{augmented} experiments we choose \( r = 20 \), meaning that we augment each reference sentence with \( 19 \) synthetic sentences.
This is chosen by manually evaluating the quality of augmented sentences on examples, as shown in Appendix~\ref{app:examples}.
We run two \texttt{randompad} experiments, \texttt{randompad(2)} (half of sentences are random), and \texttt{randompad(20)} (same ratio as \texttt{augmented(20)} for comparison).
The \texttt{half(20)} experiment has the same ratio as \texttt{augmented(20)}, with each reference sentence accompanied by \( 9 \) random sentences and \( 10 \) synthetic sentences.\footnote{
We leave it to future work to explore additional padding and augmentation ratios.
In the \texttt{half} experiments, we randomly select half of the sentences from the synthetic sentences from \texttt{augmented}: this means in particular that we do not necessarily keep the best half of the synthetic sentences, leaving room for improvement in future experiments.}

\subsection{Evaluation metrics} \label{sec:eval}
We use two evaluation metrics: (i)~the BLEU score \citep{papineni-etal-2002-bleu} computed with the \texttt{multi-bleu-detok} script from the Moses toolkit~\cite{koehn-etal-2007-moses} on the entire test set to evaluate overall MT quality, and (ii)~the accuracy of the filtered evaluation words to evaluate how well the approaches learn from the post-edits.
We use a clipped bag-of-words accuracy defined as follows: if in the target language the reference sentence contains the word \( w \) \( n \) times, and the translated sentence contains it \( p \) times, then the accuracy is \( \frac{\min(p, n)}{n} \).
This accuracy is computed separately for each of the evaluation words \( w \), and then averaged to obtain an overall accuracy.
The advantage of this metric is that it also only requires segmentation to be done in the target language, meaning that once more no prior knowledge or training further than the base model are required for the source language.

A third evaluation metric, which is very important for our scenario but overlooked in previous work, is the speed at which the model improves its accuracy.
We define this as the number of reference sentences per novel word it needs to see in order to substantially improve the accuracy in translating these words.
For a journalist correcting an MT system's mistakes, it is important to correct a given mistake as few times as possible, since having to correct each mistake beyond a certain number of times might make it more worthwhile to simply manually translate the article.
The best way of evaluating this metric is by comparing the evolution of the accuracy curve and the BLEU curve as a function of the number of seen occurrences, as presented in the next section in more detail.

We also compare all models to the reference model \texttt{all-data}, trained on the unfiltered training set, as in the University of Edinburgh's WMT19 submission~\citep{bawden-etal-2019-university}.
This model is considered as a topline rather than a baseline since it does not perform the same task and was trained with much more data -- our baseline being the \texttt{finetune} approach.

% RESULTS FIGURE
\begin{figure*}[ht]
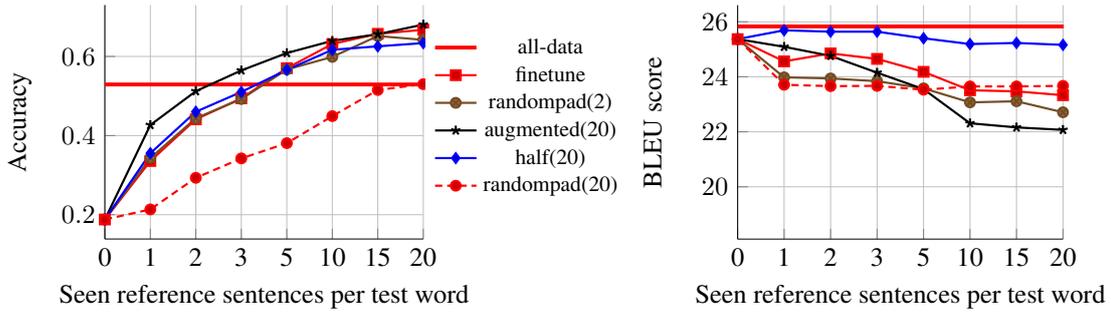

% SLOW
\centering
\resizebox{0.9\textwidth}{!}{
\begin{subfigure}{.57\textwidth}
  \include{Figures/results.slow.accuracy}
\end{subfigure}
\begin{subfigure}{.41\textwidth}
  \include{Figures/results.slow.bleu}  
\end{subfigure}}
\caption{BLEU and accuracy results of all of our approaches for the slow speed setting}
\label{fig:results-slow}
\end{figure*}

\begin{figure*}
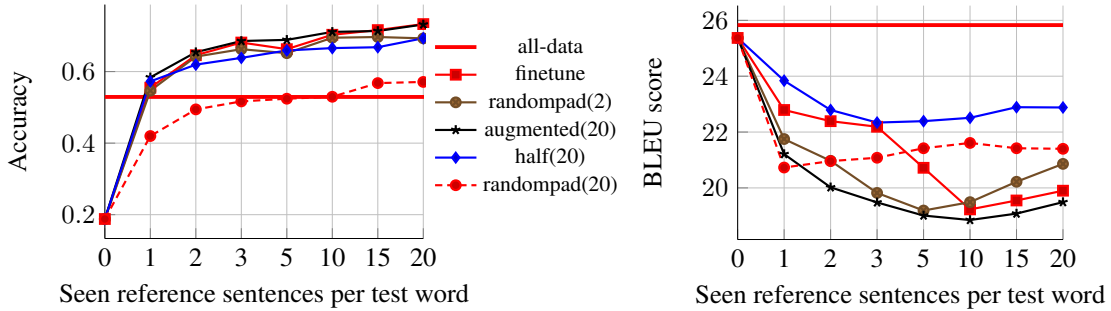

% FAST
\centering
\resizebox{0.9\textwidth}{!}{
\begin{subfigure}{.57\textwidth}
  \include{Figures/results.fast.accuracy}
\end{subfigure}
\begin{subfigure}{.41\textwidth}
  \include{Figures/results.fast.bleu}
\end{subfigure}}
\caption{BLEU and accuracy results of all of our approaches for the fast speed setting}
\label{fig:results-fast}
\end{figure*}

\subsection{Hyper-parameter choice}
To ensure optimal settings, for each approach we choose
hyper-parameters based on the size of the fine-tuning corpus.
Several hyper-parameter choices seem relevant, depending on the goal: more training leads to lower BLEU scores (as overfitting occurs) but higher accuracies on the evaluation words,
exposing a trade-off between BLEU and translation performance specifically on the
evaluation words.
This is shown in Appendix~\ref{app:tradeoff} where we explore
different values
for the \texttt{finetune} approach.
In our results in the following section, we use hyper-parameter values
that best match the accuracy scores for all approaches, thus simplifying the trade-off to a direct comparison in BLEU score.

Choosing a single epoch and learning rate value depends on the end goal:
we can focus on accuracy at all costs, even if it means decreasing overall translation performance, or take a more conservative approach by moderately increasing accuracy while maintaining BLEU.
Another way of seeing this trade-off is with respect to a time scale over which adaptation occurs: if the process is to be repeated many times, then it may be wise to decrease BLEU as little as possible, whereas a system that is often reset or used less often can afford to sacrifice more BLEU for extra accuracy on novel words.
It is important to note that these results are batched, so the BLEU losses or gains illustrated correspond to the learning of 96 words at once.

We present our results with two adaptation speeds for each approach: a slow (\textit{S}) and a fast (\textit{F}) setting, corresponding to two possible compromises in the previously explained trade-off. 
At each adaptation speed, for each approach, different hyper-parameters are used in order to match accuracy scores as closely as possible to make their comparison easier.
The table of chosen epoch and learning rate values can be found in Appendix~\ref{app:lr-epoch}.

\section{Results}\label{sec:results}
Figures~\ref{fig:results-slow} and~\ref{fig:results-fast} present the results for the reference model and all approaches for both slow and fast adaptation speeds.
We immediately notice that at both adaptation speeds our models are capable of surpassing the \texttt{all-data} model\footnote{Note however that \texttt{all-data} does not constitute a baseline for our model: it does not perform the same task and has access to many more reference sentence pairs.} in terms of accuracy on the new words, despite having seen fewer than 20 (and as few as 1) reference sentences containing those words, whereas \texttt{all-data} has seen each evaluation word over 300 times on average.
Our baseline, the \texttt{finetune} model, is shown to be largely surpassed by all other approaches in accuracy, both on the slow and fast settings.

With hyper-parameters being chosen to approximately match accuracy curves, the main comparison point is the BLEU score.
In Figure~\ref{fig:results-slow}, we see that in the slow setting our \texttt{half} approach is the only one able to improve the BLEU score while learning the new words, nearly matching \texttt{all-data} in both BLEU and accuracy at 3 occurrences seen.
The \texttt{augmented} approach offers a slightly better accuracy curve than all other models on this speed, but loses out on BLEU score at higher occurrence numbers.

It is very important to look at the first points on the curve: 1-5 occurrences of each evaluation word is the realistic range to imagine a journalist making corrections, since a journalist could become frustrated with a model requiring each new word to be corrected up to 20 times for it to be correctly translated.
In this respect, Figure~\ref{fig:results-fast} for the fast setting offers the best results, with accuracy immediately surpassing \texttt{all-data} for all but one model, albeit at a heightened cost to BLEU.
Several models, including our new \texttt{augmented} and \texttt{half} approaches, see an increase in BLEU at higher occurrence numbers, suggesting that generalization is occurring as more sentences are learned with bigger learning steps.

The \texttt{randompad} approaches surprised us in two ways, especially given \texttt{half}'s success: they generally perform worse than \texttt{finetune} despite being designed as an improvement over it, and \texttt{randompad(20)} has very poor accuracy scores compared to all other approaches.
To the first point, one possible explanation is that overfitting might occur with both the random sentences and the reference sentences when few data is used for fine-tuning (low seen occurrence numbers).
This would explain why in Figure~\ref{fig:results-fast} the BLEU curve for \texttt{randompad} ends up overtaking the \texttt{finetune} curve once enough padding data is available to allow generalization.
The second point can be explained by a form of `dilution' of the reference sentences containing the evaluation words: the model over-adapts to the sentences provided to it but does not particularly improve on the evaluation words.
We also tried various hyper-parameters for \texttt{randompad(20)} but were unable to find a compromise similar to other models, resulting in this approach standing out from the others.
This might also be partially explained through variations inherent to the randomness in the selection of the padding sentences.

\subsection{Analysis}
To gain more insight into which words were translated well or poorly, we chose to look at our accuracy metric for each evaluation word.
For the fast \texttt{half} approach, averaged across all occurrence numbers, the best words were generally proper nouns such as \textit{Sulawesi} (1.00), \textit{Isabel} (0.98) and \textit{Kohli} (0.68).
Acronyms such as \textit{ATM} (1.00), \textit{RCN} (0.98) or \textit{GB} (0.87) also performed very well, and common nouns or verbs had more varied performance: \textit{niece} (0.00), \textit{moustache} (0.37) and \textit{smartphone} (0.87).
Although the very best words are generally proper nouns or acronyms and the very worst generally other parts of speech, no clear pattern or general rule can be ascertained.

One phenomenon initially worrying us was the over-translation of evaluation words: a model outputting evaluation words where they should not appear would be able to `trick' our bag-of-words accuracy metric by artificially inflating it to a certain point.
However, we hypothesized and then verified that this would be counteracted by a decrease in BLEU score due to these words appearing in sentences where they should not.
This is presented in Appendix~\ref{app:overtrans} where we confirm this by demonstrating a negative correlation between over-translation and BLEU score.

\section{Conclusion}
We explored different techniques based on fine-tuning in order to adapt a base model to post-edits containing novel vocabulary.
We proposed a data augmentation technique never applied to this task, allowing us to expand the number of occurrences of the new words available to our model to learn from.
In our experiments, all proposed adaptation techniques offer better performance on the novel words than our reference model, which had seen the words hundreds of times each.
Our data augmentation approaches yield faster adaptation than our baseline, but with a greatly improved BLEU score, especially when combined with generalization using padding with random training set data.

These various techniques could all be applied to lifelong adaptation of an MT system often confronted with new vocabulary or expressions.
Our work can be generalized in several directions which we chose to leave for future work: word translations can automatically be retrieved with alignments, rare multi-word expressions (\textit{n-grams}) can be used rather than rare single-token words, the language model generalization can be used in the source language or both languages and several word substitutions can be made in a single sentence.
Several approaches can be deployed to further improve the BLEU score when fine-tuning, such as regularization techniques as explored by~\citet{simianer-etal-2019-measuring}, keeping only the best half of synthetic sentences rather than a random half, and a more careful choice of the ratio of augmented and random data relative to reference sentences.

\section*{Acknowledgments}
This work was supported by funding from the European Union’s Horizon 2020 research and innovation programme under grant agreement No 825299 (GoURMET) and funding by the UK Engineering and Physical Sciences Research Council (EPSRC) fellowship grant EP/S001271/1 (MTStretch).

\bibliography{anthology,emnlp2020}
\bibliographystyle{acl_natbib}

\clearpage
\appendix
\section{Filtered words}
\label{app:filtered}
The complete list of the 96 filtered words in English is given in Table~\ref{tab:words}.

\begin{figure*}[ht]
\centering\small
  \begin{tabular}{ | l | l | l | l | l | l | l | }
    \hline
2018 & ATM & Ahmedabad & Ambani & Amul & Anand & Ayr \\ \hline
BJP & Bachchan & Becker & Bedford & Chequers & Constantinople & Conway \\ \hline
DM & Dinesh & Dragons & Fidelity & Fleetwood & GB & GST \\ \hline
Gadkari & Giga & HDFC & Hastings & Isabel & Jammu & Kapoor \\ \hline
Kavanaugh & Keyser & Kohli & Lavrov & Lina & Lucknow & MLA \\ \hline
Manish & Mayorga & Meng & Modi & Molinari & Mukesh & Musk \\ \hline
Márquez & Nana & Narendra & Nifty & Oldham & Palu & Patriarch \\ \hline
Patriarchate & Prithvi & Pune & RCN & RTI & Rajkot & Rupani \\ \hline
Rupee & Sachin & Salman & Scalia & Seeley & Sensex & Shetty \\ \hline
Shilpa & Spiegel & Sulawesi & Surat & Sushma & Tendulkar & Tesla \\ \hline
Tiwari & Twitter & Vadodara & Virat & Vyas & Watts & app \\ \hline
apps & cleanliness & crores & cylinders & dough & fortress & ghee \\ \hline
inaugurate & intoxicated & lakhs & litre & mentioning & moustache & niece \\ \hline
refrigerators & sacrificed & slab & smartphone & strawberries &  & \\ \hline
  \end{tabular}
  \caption{Filtered words from the corpus} \label{tab:words}
\end{figure*}

\section{Data augmentation example}
\label{app:examples}
This is an actual example of contextual data augmentation from a given sentence.
The reference sentence is, ``\textit{A powerful 7.5 magnitude earthquake hit the Indonesian island of \textbf{Sulawesi} on Friday, September 29, triggering a tsunami and leaving nearly 400 people dead.}'', with the novel word being \textit{Sulawesi}.
The top five sentences matching the context of \textit{Sulawesi} above are as follows:
\begin{enumerate}
    \item This labour shortage prompted the authorities to import slaves from \textbf{Indonesia} and Madagascar.
    \item Many of them have settled down in Ahmedabad, Vadodara, Mumbai, Kolkota, Delhi, Nagpur and far away places like \textbf{Java}, Rangoon, Singapore, Fiji, Eden, Kenya, Uganda, America etc and established their business in these places.
    \item The rice lands of \textbf{Java} are among the richest in the world.
    \item Rising ocean temperatures and ocean acidification means that the capacity of the ocean carbon sink will gradually get weaker, giving rise to global concerns expressed in the \textbf{Monaco} and Manado Declarations.
    \item \textbf{Lara}'s first school was St. Joseph's Roman Catholic primary.
\end{enumerate}

These sentences generally capture the idea that the word refers to an island, except for the last one which might be an outlier.
The fourth sentence is remarkable in the fact that \textit{Monaco} is not an island, but the context of \textit{rising ocean temperatures and ocean acidification} as well as the mention of \textit{Manado} alongside (the capital of North Sulawesi) make the sentence relevant here.
This highlights the importance of context besides the actual words' meanings.

\section{Learning rate and epochs}
\label{app:lr-epoch}
Table~\ref{tab:lr-epoch} gives the epoch numbers and learning rates we chose for our presented examples.
The \texttt{augmented} and \texttt{half} experiments both have lower learning rates because they fine-tune on more data, and thus go through more gradient steps.

\begin{figure*}[ht]
\centering\small
  \begin{tabular}{llrrrrr}
    \toprule
& & \texttt{finetune} & \texttt{randompad(2)} & \texttt{randompad(20)} & \texttt{augmented} & \texttt{half} \\ 
\midrule
\multirow{2}{*}{Slow (S)} & Epochs & 10 & 10 & 10 & 10 & 10 \\ 
 & Learning rate& \(4 \times 10^{-5}\) & \(4 \times 10^{-5}\) & \(1 \times 10^{-5} \) & \(4 \times 10^{-6}\) & \(4 \times 10^{-6}\) \\
 \midrule
\multirow{2}{*}{Fast (F)} & Epochs & 30 & 30 & 30 & 10 & 10 \\
& Learning rate & \(1 \times 10^{-4}\) & \(1 \times 10^{-4}\) & \(4 \times 10^{-5}\) & \(4 \times 10^{-5}\) & \(4 \times 10^{-5}\) \\
\bottomrule
  \end{tabular}
  \caption{The chosen learning rate and epoch values in our experiments} \label{tab:lr-epoch}
\end{figure*}

\section{Preliminary survey of hyper-parameters}
\label{app:tradeoff}
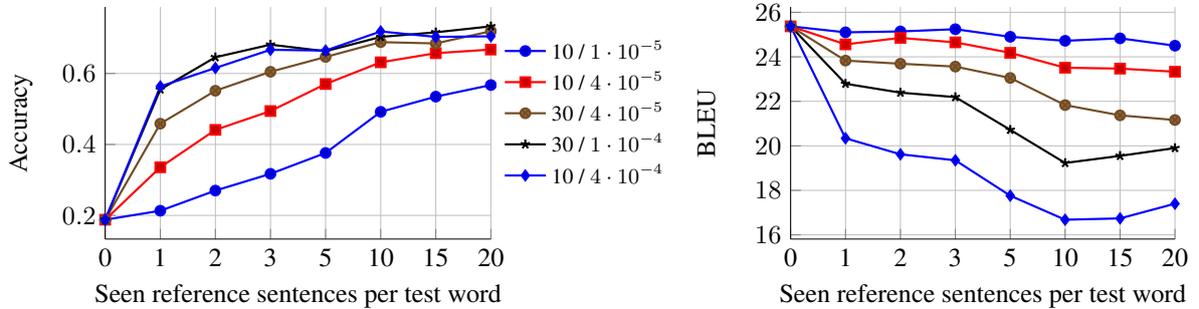
\begin{figure*}[ht]
\resizebox{\linewidth}{!}{
  \input{Figures/lr.epoch.calibration.accuracy}
  \input{Figures/lr.epoch.calibration.bleu}
}
\caption{Evolution of the \texttt{finetune} approach's performance labeled by \textit{epochs} / \textit{learning rate}}
\label{fig:lr-epoch}
\end{figure*}

We tried different hyper-parameter combinations for the \texttt{finetune} approach to explore the dependency of accuracy and BLEU score to these.
Figure~\ref{fig:lr-epoch} shows these curves labeled by `\textit{number of epochs} / \textit{learning rate}'.

We noticed a clear trade-off between BLEU score and accuracy as the parameters evolved, with more training leading to bigger gains in accuracy at a cost to BLEU.
Higher parameters saw diminishing returns as accuracy would improve little or not at all while BLEU kept decreasing, as can be seen for the two first curves.
Lower parameters also showed diminishing returns in BLEU score as accuracy dramatically drops.

% LR EPOCH FIGURE

\section{Over-translation}
\label{app:overtrans}
Table~\ref{tab:overtrans} shows the over-translation metric for different approaches. Over-translation is defined as the number of times each word appears in sentences in excess of the reference sentence, divided by the number of times it appears in the reference.
This metric is computed similarly to the accuracy metric as explained in Section~\ref{sec:eval}, i.e.~per word over all sentences and then averaged over all words.

For reference, the \texttt{all-data} model gets an over-translation metric of \( 0.04 \) and the base model (the one referred to as 0 occurrences in figures) scores \( 0.01 \).

While some approaches have seemingly high over-translation values, the most competitive approaches do not see such a high increase in over-translation.
For instance, the over-translation metric of \( 0.25 \) for \texttt{half(20)} (F) means that an over-translated word would appear for every four reference occurrences of our evaluation words, which means that it would over-translate only in about one in seventeen sentences in the test set (given that reference words only appear in \( 23 \% \) of test set sentences).
The \texttt{half} approach has very good BLEU scores, which is aligned with the fact that it has some of the best over-translation scores amongst the presented approaches.

\begin{figure*}[ht]
\centering
\includegraphics[height=7cm]{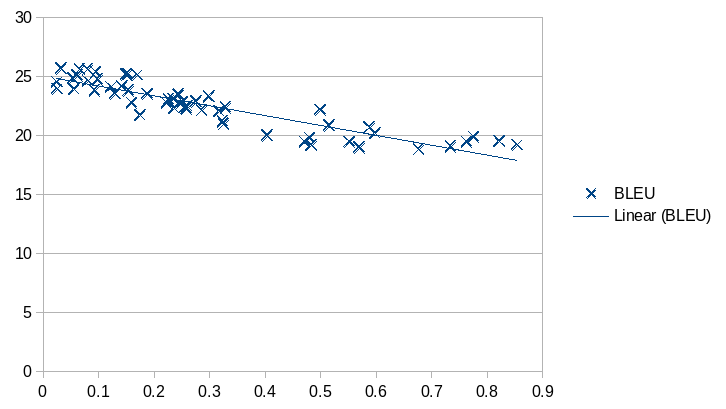}
\caption{BLEU score as a function of over-translation score}
\label{fig:overtrans}
\end{figure*}

Moreover, Figure~\ref{fig:overtrans} shows for all of these data points the evolution of BLEU score with over-translation: there is a clear linear correlation, implying that BLEU captures over-translation by our approaches very well.

\begin{figure*}[ht]
\centering\small
  \begin{tabular}{r | r r r r r r r}
\toprule
Seen reference occurrences & 1 & 2 & 3 & 5 & 10 & 15 & 20 \\
\hline
\texttt{finetune} (S) & 0.02 & 0.05 & 0.08 & 0.14 & 0.24 & 0.24 & 0.3 \\
\texttt{finetune} (F) & 0.16 & 0.33 & 0.5  & 0.59 & 0.85 & 0.82 & 0.78 \\
\texttt{randompad(2)} (S) & 0.03 & 0.06 & 0.09 & 0.13 & 0.23 & 0.23 & 0.25 \\
\texttt{randompad(2)} (F) & 0.17 & 0.33 & 0.48 & 0.48 & 0.55 & 0.6 & 0.52 \\
\texttt{augmented(20)} (S) & 0.06 & 0.1 & 0.12 & 0.19 & 0.26 & 0.29 & 0.32 \\
\texttt{augmented(20)} (F) & 0.32 & 0.4 & 0.47 & 0.57 & 0.68 & 0.73 & 0.76 \\
\texttt{half(20)} (S) & 0.03 & 0.07 & 0.08 & 0.1 & 0.15 & 0.15 & 0.17 \\
\texttt{half(20)} (F) & 0.15 & 0.22 & 0.24 & 0.26 & 0.26 & 0.28 & 0.25 \\
\bottomrule
  \end{tabular}
  \caption{The over-translation score of various approaches} \label{tab:overtrans}
\end{figure*}

\end{document}

%% file: Figures/results.slow.accuracy.tex
\begin{tikzpicture}

	\begin{axis}[
		height=5cm,
		width=6.2cm,
		xmin = 0,
		xmax = 7,
		grid=major,
		ylabel = Accuracy,
		xlabel = Seen reference sentences per test word,
		xtick = {0,1,2,3,4,5,6,7},
		xticklabels = {0,1,2,3,5,10,15,20},
        axis x line*=bottom,
        axis y line*=left,
        legend style={draw=none,
            at={(1.65, .9), font=\small,
            /tikz/every even column/.append style={column sep=0.9cm}},
    },
    legend columns=1
	]
\addlegendentry{all-data}
\addplot[mark=none, ultra thick, color=red] coordinates {
 (0, 0.5291) (20, 0.5291)};
 
  \addlegendentry{finetune} % 10/00004
\addplot coordinates {
 (0, 0.187893469717918) (1, 0.335693519291865) (2, 0.441051179136132) (3, 0.494112688624451) (4, 0.570300734986645) (5, 0.631444639096812) (6, 0.657169711717893) (7, 0.667387755416766)};
 
  \addlegendentry{randompad(2)} % 10/00004
\addplot coordinates {
 (0, 0.187893469717918) (1, 0.343187348973195) (2, 0.443857903974107) (3, 0.492334652940165) (4, 0.5670035356515) (5, 0.598870628934691) (6, 0.651631332722318) (7, 0.641872616835057)};
 
 \addlegendentry{augmented(20)} % 10/000004
\addplot coordinates {
 (0, 0.187893469717918) (1, 0.426679716685625) (2, 0.512160743801852) (3, 0.564632661895246) (4, 0.608582633098507) (5, 0.639444818438764) (6, 0.656198228184559) (7, 0.680521996469067)};
 
  \addlegendentry{half(20)} % 10/000004
\addplot coordinates {
 (0, 0.187893469717918) (1, 0.355327965356195) (2, 0.460187667349356) (3, 0.510293584054054) (4, 0.56606556059377) (5, 0.616823269363697) (6, 0.625535980500909) (7, 0.633726559179933)};
 
  \addlegendentry{randompad(20)} % 10/00001
\addplot coordinates {
 (0, 0.187893469717918) (1, 0.213390301657765) (2, 0.293228962319692) (3, 0.342309437696124) (4, 0.380839138823058) (5, 0.449149565898895) (6, 0.515194757097849) (7, 0.529913207007206)};
 
\end{axis}
\end{tikzpicture}

%% file: Figures/results.slow.bleu.tex
\begin{tikzpicture}
	\begin{axis}[
		height=5cm,
		width=6.3cm,
		grid=major,
		xmin=0,
		xmax=7,
		ymin=18.1,
		ylabel = BLEU score,
		xlabel = Seen reference sentences per test word,
		xtick = {0,1,2,3,4,5,6,7},
		xticklabels = {0,1,2,3,5,10,15,20},
        axis x line*=bottom,
        axis y line*=left
	]
	
\addlegendentry{all-data}
\addplot[mark=none, ultra thick, color=red] coordinates {
 (0, 25.83) (20, 25.83)};
 
\addlegendentry{finetune (S)} % 10/00004
\addplot coordinates {
 (0, 25.37) (1, 24.56) (2, 24.85) (3, 24.65) (4, 24.18) (5, 23.51) (6, 23.47) (7, 23.33)};
 
 \addlegendentry{randompad (S)} % 10/00004
\addplot coordinates {
 (0, 25.37) (1, 23.98) (2, 23.94) (3, 23.84) (4, 23.58) (5, 23.07) (6, 23.11) (7, 22.71)};
 
\addlegendentry{augmented (S)} % 10/000004
\addplot coordinates {
 (0, 25.37) (1, 25.09) (2, 24.76) (3, 24.15) (4, 23.55) (5, 22.31) (6, 22.16) (7, 22.07)};
 
 \addlegendentry{half (S)} % 10/000004
\addplot coordinates {
 (0, 25.37) (1, 25.69) (2, 25.64) (3, 25.64) (4, 25.4) (5, 25.19) (6, 25.23) (7, 25.16)};
 
  \addlegendentry{randomBig (S)} % 10/00001
\addplot coordinates {
 (0, 25.37) (1, 23.71) (2, 23.66) (3, 23.67) (4, 23.53) (5, 23.65) (6, 23.65) (7, 23.67)};

\legend{}
\end{axis}
\end{tikzpicture}

%% file: Figures/results.fast.accuracy.tex
\begin{tikzpicture}

	\begin{axis}[
		height=5cm,
		width=6.2cm,
		grid=major,
		xmin=0,
		xmax=7,
		ylabel = Accuracy,
		xlabel = Seen reference sentences per test word,
		xtick = {0,1,2,3,4,5,6,7},
		xticklabels = {0,1,2,3,5,10,15,20},
        axis x line*=bottom,
        axis y line*=left,
        legend style={draw=none,
            at={(1.65, .9), font=\small,
            /tikz/every even column/.append style={column sep=0.9cm}},
    },
    legend columns=1
	]
	
\addlegendentry{all-data}
\addplot[mark=none, ultra thick, color=red] coordinates {
 (0, 0.5291) (20, 0.5291)};
	
\addlegendentry{finetune}
\addplot coordinates {
 (0, 0.187893469717918) (1, 0.555018575070751) (2, 0.645757751796458) (3, 0.681216796806725) (4, 0.662687225050532) (5, 0.702953089554842) (6, 0.716000763855799) (7, 0.732807029183079)};

 \addlegendentry{randompad(2)} % 30/0001
\addplot coordinates {
 (0, 0.187893469717918) (1, 0.546560614618546) (2, 0.641826176788479) (3, 0.662580080064842) (4, 0.651456281319588) (5, 0.69465385013231) (6, 0.696619828595932) (7, 0.692862571692528)};
 
%\addlegendentry{augmented (M)} % 10/00001
%\addplot coordinates {
% (0, 0.187893469717918) (1, 0.519120114829148) (2, 0.597340204887957) (3, 0.616579847563925) (4, 0.653984782988439) (5, 0.683416846527645) (6, 0.690772396653678) (7, 0.707374903186594)};
 
 \addlegendentry{augmented(20)} % 10/00004
\addplot coordinates {
 (0, 0.187893469717918) (1, 0.583865935469139) (2, 0.654156503731982) (3, 0.685478179438644) (4, 0.688639477878959) (5, 0.710919582889362) (6, 0.713476478792123) (7, 0.731613902777761)};
 
  \addlegendentry{half(20)} % 10/00004
\addplot coordinates {
 (0, 0.187893469717918) (1, 0.572552966820955) (2, 0.619449401444267) (3, 0.638312897284522) (4, 0.659073469175989) (5, 0.665648928973094) (6, 0.668092690831523) (7, 0.693172923948645)};
 
  \addlegendentry{randompad(20)} % 30/00004
\addplot coordinates {
 (0, 0.187893469717918) (1, 0.419832307539109) (2, 0.494310936661321) (3, 0.516632670334987) (4, 0.524174802006583) (5, 0.529620351953978) (6, 0.567820599229994) (7, 0.571085263420997)};
 
\end{axis}
\end{tikzpicture}

%% file: Figures/results.fast.bleu.tex
\begin{tikzpicture}
	\begin{axis}[
		height=5cm,
		width=6.3cm,
		grid=major,
		xmin=0,
		xmax=7,
		ylabel = BLEU score,
		xlabel = Seen reference sentences per test word,
		xtick = {0,1,2,3,4,5,6,7},
		xticklabels = {0,1,2,3,5,10,15,20},
        axis x line*=bottom,
        axis y line*=left
	]
	
\addlegendentry{all-data}
\addplot[mark=none, ultra thick, color=red] coordinates {
 (0, 25.83) (20, 25.83)};
 
\addlegendentry{finetune (F)} % 30/0001
\addplot coordinates {
 (0, 25.37) (1, 22.79) (2, 22.39) (3, 22.19) (4, 20.72) (5, 19.23) (6, 19.55) (7, 19.9)};
 
 \addlegendentry{randompad (F)} % 30/0001
\addplot coordinates {
 (0, 25.37) (1, 21.75) (2, 20.97) (3, 19.82) (4, 19.19) (5, 19.49) (6, 20.22) (7, 20.86)};
  
% \addlegendentry{augmented (M)} % 10/00001
%\addplot coordinates {
% (0, 25.37) (1, 24.11) (2, 23.27) (3, 22.47) (4, 21.81) (5, 21.09) (6, 20.82) (7, 20.96)};
 
 \addlegendentry{augmented (F)} % 10/00004
\addplot coordinates {
 (0, 25.37) (1, 21.21) (2, 20.02) (3, 19.48) (4, 19.01) (5, 18.85) (6, 19.08) (7, 19.49)};
 
  \addlegendentry{half (F)} % 10/00004
\addplot coordinates {
 (0, 25.37) (1, 23.84) (2, 22.79) (3, 22.34) (4, 22.39) (5, 22.51) (6, 22.89) (7, 22.88)};
 
 \addlegendentry{randomBig (F)} % 30/00004
\addplot coordinates {
 (0, 25.37) (1, 20.73) (2, 20.96) (3, 21.08) (4, 21.42) (5, 21.61) (6, 21.42) (7, 21.4)};

\legend{}
\end{axis}
\end{tikzpicture}

%% file: Figures/lr.epoch.calibration.accuracy.tex
\begin{tikzpicture}

	\begin{axis}[
		height=5cm,
		width=7.22cm,
		grid=major,
		xmin=0,
		xmax=7,
		ylabel = Accuracy,
		xlabel = Seen reference sentences per test word,
		xtick = {0,1,2,3,4,5,6,7},
		xticklabels = {0,1,2,3,5,10,15,20},
        axis x line*=bottom,
        axis y line*=left,
        legend style={draw=none,
            at={(1.48, .9), font=\small,
            /tikz/every even column/.append style={column sep=0.9cm}},
    },
    legend columns=1
	]
% \addlegendentry{all-data}
% \addplot[mark=none, ultra thick, color=red] coordinates {
% (0, 0.5291) (20, 0.5291)};

  \addlegendentry{10 / \(1 \cdot 10^{-5}\)}
\addplot coordinates {
 (0, 0.187893469717918) (1, 0.213225245784988) (2, 0.269991929196892) (3, 0.317259335771022) (4, 0.376031759114362) (5, 0.491927548384212) (6, 0.534713283927957) (7, 0.567361140504385)};
 
   \addlegendentry{10 / \(4 \cdot 10^{-5}\)}
\addplot coordinates {
 (0, 0.187893469717918) (1, 0.335693519291865) (2, 0.441051179136132) (3, 0.494112688624451) (4, 0.570300734986645) (5, 0.631444639096812) (6, 0.657169711717893) (7, 0.667387755416766)};
 
    \addlegendentry{30 / \(4 \cdot 10^{-5}\)}
\addplot coordinates {
 (0, 0.187893469717918) (1, 0.458522210792431) (2, 0.55134312862216) (3, 0.604765818767125) (4, 0.64656439723981) (5, 0.688300449458399) (6, 0.684649553209164) (7, 0.719053595079586)};
 
     \addlegendentry{30 / \(1 \cdot 10^{-4}\)}
\addplot coordinates {
 (0, 0.187893469717918) (1, 0.555018575070751) (2, 0.645757751796458) (3, 0.681216796806725) (4, 0.662687225050532) (5, 0.702953089554842) (6, 0.716000763855799) (7, 0.732807029183079)};
 
      \addlegendentry{10 / \(4 \cdot 10^{-4}\)}
\addplot coordinates {
 (0, 0.187893469717918) (1, 0.5634490563525) (2, 0.615512643838848) (3, 0.66670656831651) (4, 0.664556743958661) (5, 0.718412647983151) (6, 0.703085498808838) (7, 0.704679547947668)};
 
\end{axis}
\end{tikzpicture}

%% file: Figures/lr.epoch.calibration.bleu.tex
\begin{tikzpicture}
	\begin{axis}[
		height=5cm,
		width=7.2cm,
		grid=major,
		xmin=0,
		xmax=7,
		ylabel = BLEU,
		xlabel = Seen reference sentences per test word,
		xtick = {0,1,2,3,4,5,6,7},
		xticklabels = {0,1,2,3,5,10,15,20},
        axis x line*=bottom,
        axis y line*=left
	]
	
%\addlegendentry{all-data}
%\addplot[mark=none, ultra thick, color=red] coordinates {
% (0, 25.83) (20, 25.83)};
 
  \addlegendentry{10 / \(1 \cdot 10^{-5}\)}
\addplot coordinates {
 (0, 25.37) (1, 25.1) (2, 25.14) (3, 25.24) (4, 24.9) (5, 24.72) (6, 24.83) (7, 24.5)};
 
   \addlegendentry{10 / \(4 \cdot 10^{-5}\)}
\addplot coordinates {
 (0, 25.37) (1, 24.56) (2, 24.85) (3, 24.65) (4, 24.18) (5, 23.51) (6, 23.47) (7, 23.33)};
 
    \addlegendentry{30 / \(4 \cdot 10^{-5}\)}
\addplot coordinates {
 (0, 25.37) (1, 23.83) (2, 23.69) (3, 23.56) (4, 23.05) (5, 21.83) (6, 21.37) (7, 21.16)};
 
     \addlegendentry{30 / \(1 \cdot 10^{-4}\)}
\addplot coordinates {
 (0, 25.37) (1, 22.79) (2, 22.39) (3, 22.19) (4, 20.72) (5, 19.23) (6, 19.55) (7, 19.9)};
 
      \addlegendentry{10 / \(4 \cdot 10^{-4}\)}
\addplot coordinates {
 (0, 25.37) (1, 20.34) (2, 19.62) (3, 19.35) (4, 17.76) (5, 16.68) (6, 16.74) (7, 17.4)};
 
\legend{}
\end{axis}
\end{tikzpicture}